\documentclass[runningheads]{llncs}
\usepackage[T1]{fontenc}
\usepackage{caption}
\usepackage{subcaption}

\usepackage{graphicx}
\begin{document}
\title{SemUV: Deep Learning based semantic manipulation over UV texture map of virtual human heads}

%
\titlerunning{SemUV}
\authorrunning{A. Mukherjee et al.}
%

\author{Anirban Mukherjee \inst{1}\orcidID{0000-0003-3404-7045} \and
Venkat Suprabath Bitra\inst{1} \and
Vignesh Bondugula \inst{1} \and
Tarun Reddy Tallapureddy \inst{1} \and
Dinesh Babu Jayagopi \inst{1}\orcidID{0000-0003-0080-452X}}

\institute{International Institute of Information Technology Bangalore}

\maketitle              
\begin{abstract}
Designing and manipulating virtual human heads is essential across various applications, including AR, VR, gaming, human-computer interaction and VFX. Traditional graphic-based approaches require manual effort and resources to achieve accurate representation of human heads. While modern deep learning techniques can generate and edit highly photorealistic images of faces, their focus remains predominantly on 2D facial images. This limitation makes them less suitable for 3D applications. Recognizing the vital role of editing within the UV texture space as a key component in the 3D graphics pipeline, our work focuses on this aspect to benefit graphic designers by providing enhanced control and precision in appearance manipulation. Research on existing methods within the UV texture space is limited, complex, and poses challenges. In this paper, we introduce SemUV: a simple and effective approach using the FFHQ-UV dataset for semantic manipulation directly within the UV texture space. We train a StyleGAN model on the publicly available FFHQ-UV dataset, and subsequently train a boundary for interpolation and semantic feature manipulation. Through experiments comparing our method with 2D manipulation technique, we demonstrate its superior ability to preserve identity while effectively modifying semantic features such as age, gender, and facial hair. Our approach is simple, agnostic to other 3D components such as structure, lighting, and rendering, and also enables seamless integration into standard 3D graphics pipelines without demanding extensive domain expertise, time, or resources.

\keywords{Computer Graphics  \and Computer Vision \and Deep Learning \and Generative Models \and Virtual Humans}
\end{abstract}
\section{Introduction}

\label{sec:intro}

Virtual humans are computer-generated characters designed to appear and behave like real people, created using technologies such as computer graphics, animation, and artificial intelligence. Creating and animating virtual humans is an active area of research due to their use in various application domains such as VR, AR, Mixed Reality, Human-Computer Interaction, Gaming and entertainment such as VFX and CGI. Designing and editing the facial appearance of the virtual human is a crucial and challenging task, whether the virtual human is based on an existing human, or a generated identity. The appearance comprises of semantic features, i.e. features in the image space having an interpreted meaning, such as age, expression, and facial hair, which need be considered with respect to the context and use-cases while designing and animating virtual human heads.

In traditional 3D graphics, the process of creating such virtual human heads primarily include designing a 3D mesh model, followed by designing a texture map, which is wrapped onto the mesh to give color and material information. These textures such as albedo (representing base color) and normals (representing surface depth) are 2D images, which are mapped onto the mesh using the concept of UV mapping, which helps in specifying how a 3D model's surface will be colored as per the 2D texture image. Therefore, the texture maps lie in the UV coordinates which, although being of 2-dimensions, is different from the XY coordinate of image space where regular images are represented. To design the appearance of a virtual head, graphic designers either paint on the UV space, or directly on the 3D model, which is a challenging and time-consuming task, requires manual effort and expertise in graphics, and difficult to achieve photorealistic results.

\begin{figure}[h!]
    \centering
    \includegraphics[width=0.8\textwidth]{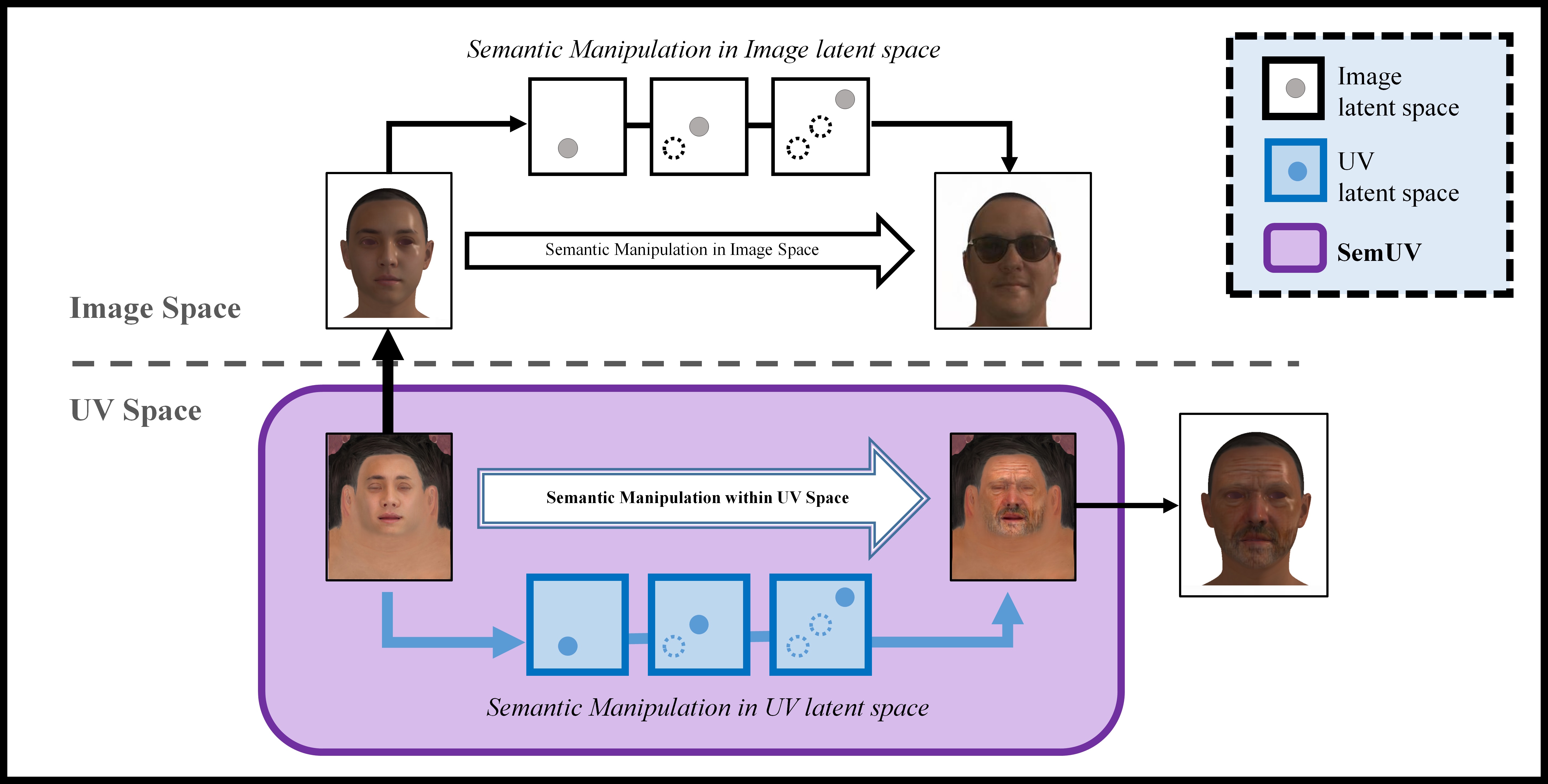}
    \caption{SemUV vs Image based approach: Our approach works only on the \textit{UV space}, focusing on the face texture, thus preventing unwanted changes possible in the image domain. Moreover, we can see the pathway for performing changes directly in UV space is shorter, thus making the approach simpler and faster.}
    \label{fig:SemUV_vs_Img}
\end{figure}

Comparatively, generating and editing regular 2D images, as compared to 3D models, have greatly simplified lately due to the latest advancements in fields of AI such as deep learning and computer vision, where deep neural networks can learn the distribution of these images from thousands of diverse examples in a high dimensional latent space, which allows a user to generate, manipulate and animate photorealistic faces \cite{karras2019style}, \cite{karras2020analyzing},  \cite{shi2022semanticstylegan}. Although 2D human images can be edited using some of these approaches, they lack explicit structural information inherent in 3D head objects. The absence of geometry requires the 2D based models to implicitly learn the 3D movements \cite{grassal2022neural}. This leads to structural and temporal consistencies and often results in artifacts while talking, performing pose changes and behavioral movements of the virtual heads. Furthermore, creating semantic manipulations in image space could lead to undesirable changes such as modifications in head structure and appearance of unwanted objects, because of the generative models' holistic learning approaches, making it difficult to disentangle specific features implicitly. For instance, while modifying the age of a young person, a generative model might introduce spectacles in the older version of the image (Fig. \ref{fig:result_UV_vs_IMG}), which, although makes sense from a logical point of view, isn't ideal for 3D modeling purposes (Fig. \ref{fig:SemUV_vs_Img}). Such generative models, capable of learning 2D images, could be useful for learning the 2D texture maps, making it bridge the gap between 2D images and 3D computer graphics. However, there is significantly less work done for editing in the texture space for 3D human models. Consequently, our work is a crucial step in advancing the field in this direction. Some works in the lines of reconstructing 3D human heads from 2D images predict the texture maps, along with a 3D mesh. However, most of these approaches do not provide the option for editing semantic features on the texture maps. Moreover, many of these works model only the face area and not the whole head. Some of the works which perform editing in the texture space are trained using inverse rendering approaches which are complicated, dependent on the mesh, and require high computation resources.

Considering the challenges and the limitations of existing approaches, we propose SemUV: a novel method of performing semantic manipulation on the UV texture space of explicitly modelled 3D virtual human heads. We use the dataset provided by FFHQ-UV \cite{Bai_2023_CVPR}, and train a generative model to learn the distribution of the texture space of 3D heads. Then, we train a boundary using the available labels for disentangling the latent space of the generative model to be able to perform linear interpolation which correspond to semantic feature manipulation in the UV texture space. Our solution is simple to train because of its ability to directly learn from texture maps without relying on the 3D mesh structure, or applying any inverse rendering approaches. Given a texture map, our approach is significantly fast because of the simple approach of linear interpolation in the learned latent space, allowing online manipulation of appearances directly in the texture space. Through comprehensive experiments and evaluations, we demonstrate the qualitative and quantitative advantages of SemUV over existing editing approach on the image-space. We discuss the usefulness and the scope of our work in the domains of 3D computer graphics and vision, and explain how SemUV could offer substantial advantages to the broader community. We also discuss the limitations of our approach and propose possible future research directions aimed at filling the existing research gaps and progress the area of 3D virtual human editing.

\noindent To summarize, our novel contributions in this work are as follows:
\begin{itemize}
    \item We propose a simple and novel deep learning based method for supervised semantic manipulation of 3D faces by manipulating their UV texture maps in 2D space.
    \item We provide a learned UV texture space and present the results of semantic manipulation of age, beard and gender along with stage parameters such as lighting, and pose using UV texture and successfully demonstrated in the realistic 3D faces from various angles.
    \item We perform a user study to understand the perceived differences between our approach and 2D image based approach, and perform statistical tests to infer insights.
    \item We discuss how successful semantic manipulation in the texture space will be beneficial in various real life applications such as VR, AR, etc, and will allow a significant reduction in the effort and resource requirements for designers of realistic virtual human.
\end{itemize}

\noindent To the best of our knowledge, our method is the simplest and  effective solution for controlled semantic manipulation on UV texture space of 3D human heads.

\section{Related works}
\label{sec:related_works}

The topic of generation and manipulation of facial images has been a common area of research in computer vision \cite{kammoun2022generative}. Modern generative models such as styleGAN \cite{karras2019style} learn the distribution of data, thereby making it possible to generate photorealistic face images with disentangled \textit{styles}. While they provide their method for mapping an image onto the learned latent space, works such as \cite{zhu2020domain} improve the process, allowing more realistic encoding of images to latent space. 

Apart from generation, manipulation of semantic features is an important area of research. In this regard, the authors of \cite{shen2020interfacegan} have achieved significant results by disentangling the latent space through learning linear subspaces. This allows step-wise manipulation of feature, allowing explicit control of the semantic features. While the 2D facial-image space has been greatly benefited from such generative vision models, they lack a 3D geometric representation making them unsuitable for structural deformations, which is inherently present in 3D computer graphics models. Lately, there has been some works which aim at representing heads in 3D using 2D images. Most notably, models such as 3DMM \cite{egger20203d} and FLAME \cite{li2017learning} are able to generate 3D colored head models, which can be rendered on 2D screen. 3DMM model represent colors using colored vertices, while some of the implementations of FLAME incorporates texture maps along with a 3D mesh. While these works can effectively generate and reconstruct human heads, none of them offer a means to manipulate their texture maps. Similarly, works such as AvatarMe++ \cite{lattas2021avatarme++}, StyleUV \cite{lee2020styleuv}, StyleFaceUV \cite{chung2022stylefaceuv} and UV-GAN \cite{deng2018uv} also have the capability to estimate texture maps. The authors of FFHQ-UV \cite{Bai_2023_CVPR} point out some of the challenges of the existing approaches, and propose a state-of-the-art pipeline for creating full head albedo UV maps from images. They publicly release the dataset, along with labels on the identities, and we use this dataset to train our GAN model, to allow us to peform semantic manipulations on extracted texture maps using this method. While these approaches are targeted towards generating UV maps, manipulating them remains unaddressed in these works. 

The authors of \cite{rai2024towards} learns to estimate 3D shape and texture. However, the pipeline works on both a mesh and a texture level, making it difficult to train a general UV based model for all faces. In contrast, our method focuses solely on the texture space, and is trained using a simpler approach that does not necessitate the use of inverse rendering. Similarly, the authors of \cite{wu2023text} have proposed a method for generating and editing faces in 3D. Compared to their diffusion \cite{ho2020denoising} \cite{zhang2023adding} based generative model, our GAN based approach is less data-hungry and easier to train for our use case. This is because to the simpler distribution of FFHQ-UV dataset used in our method, which contrasts with the complex data distributions where diffusion models typically excel. Moreover, while their text driven approach is useful for variations in the appearance, they do not discuss about a step-wise controlled manipulation of features, a crucial concern from a 3D graphic designer's perspective. Our method ensures seamless editing of texture maps in an explicitly controlled approach, thus making it compatible with existing 3D graphics pipeline.

As an alternative to explicit modelling approaches, semantic manipulation has also been studied the realm of implicit modelling methods, such as approaches based on NeRFs \cite{mildenhall2021nerf}, which learn the volumetric or geometric information within a neural network. Although there have been methods attempting to disentangle appearance features in this space \cite{hong2022headnerf} \cite{ma2022neural}, the usage of such implicit models are limited because of a lack of strongly defined explicit structure \cite{grassal2022neural}. Therefore, such implicit models remain inferior as of yet, when compared to explicit 3D models in terms of structural consistency, which is one of the crucial areas of focus in our work.

\begin{figure}[t]
    \centering
    \includegraphics[width=\textwidth]{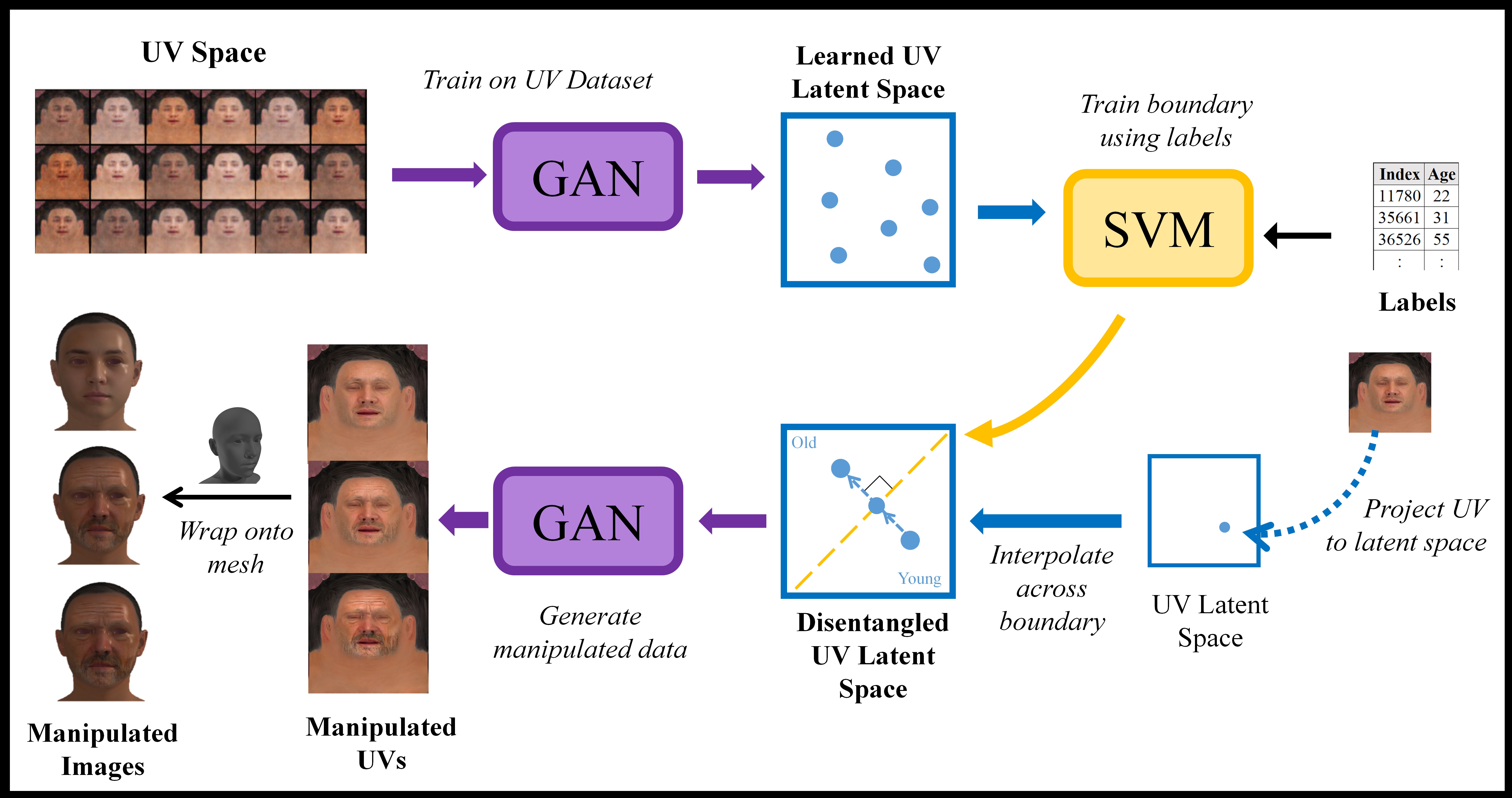}
    \caption{SemUV Overview: In our approach 1. First we learn the distribution of the UV space using a generative model, in this case StyleGAN. 2. Then, using available labels, we train a linear classifier to learn the decision boundary for the semantic features. 3. Now, given a new UV image, we project it to the UV latent space, and interpolate it in the space across the learned boundary to perform disentangled semantic manipulation. 4. Finally, we take this new latent vector and use our trained GAN to generate the semantically modified UV map, which is wrapped onto the head mesh and rendered for the final output.}
    \label{fig:SemUV_main}
\end{figure}

\section{Our approach}
\label{sec:our_approach}
In our proposed method, we operate within the space of \textit{albedo UV texture maps}, which store the fundamental color information \cite{hughes2014computer} of a virtual human head. Our method involves learning the UV texture feature space using a generative model, subsequently allowing us to perform interpolations in the latent texture space of the model, corresponding to semantic manipulations in the UV texture map. Once the texture map is wrapped onto a head mesh, the rendered results have the changes in the virtual  human's head (Fig. \ref{fig:SemUV_main}). To train such as model, first we sample from a labeled dataset \cite{Bai_2023_CVPR} with UV albedo texture maps of human faces. Then, we learn the distribution of the texture space using the architecture of StyleGANv2-ada \cite{karras2020analyzing}. After this, we train an SVM to learn a boundary for disentangling the latent space of the trained GAN model via subspace projection \cite{shen2020interfacegan}. This allows linear interpolation in the latent space resulting in precise manipulation of semantics of the face such as facial hair, age and gender without changing other aspects such as identity.

\subsection{FFHQ-UV Dataset}
We have used the FFHQ-UV dataset \cite{Bai_2023_CVPR}, which is derived from the FFHQ dataset \cite{karras2019style}, a large-scale face image dataset. FFHQ-UV contains over 50,000 images of 1024 $\times$ 1024 resolution of UV albedo maps of human faces. The dataset also has labels corresponding to the images, providing information regarding high-level semantic features such as facial hair, age and gender for each of the images. We use the images for learning the distribution of the image space of these UV texture maps, which we refer to as the UV-space. Then, we use the labels for the semantic features for training our SVM which will be used for performing the desired semantic manipulations. 
\begin{figure}[t]
    \centering
    \includegraphics[width=\textwidth]{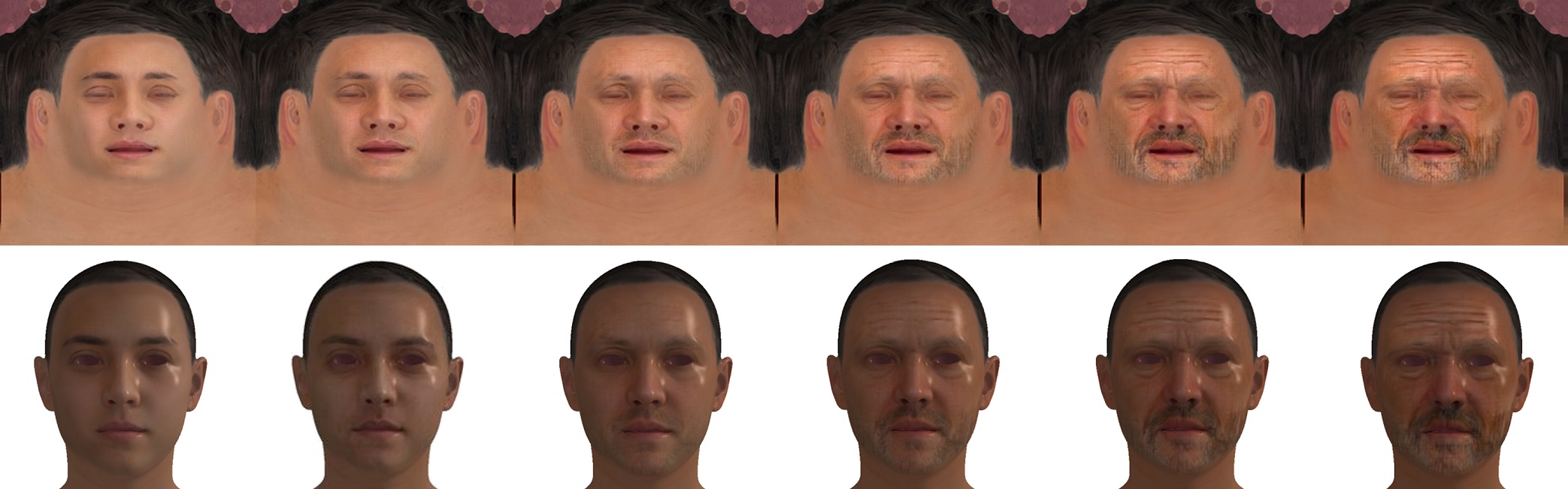}
    \caption{Outputs of SemUV: Images from \textit{left} to \textit{right} represent increasing age in the (\textit{top}) UV-Texture map and (\textit{bottom}) the final head mesh wrapped with the texture}
    \label{fig:SemUV_outputs}
\end{figure}
\subsection{Learning the UV texture space}
To learn the UV space, we train a state-of-the-art generative vision model, commonly used for facial-image generation. StyleGANv2-Ada \cite{karras2020training} is a generative adversarial network (GAN) \cite{goodfellow2014generative} based architecture, whose strengths are a combination of StyleGANv2 \cite{karras2020analyzing} and Adaptive Instance Normalization (AdaIN) \cite{huang2017arbitrary}. Similar to the basic version of StyleGAN \cite{karras2019style}, the generator produces an intermediate feature map, $w$, by mapping $z$ through a mapping network, denoted as $W$. This mapping network learns a non-linear mapping function that projects $z$ to a disentangled latent space, where different dimensions control various attributes of the generated image. The intermediate feature map $w$ is then passed through a series of convolutional layers, each followed by an AdaIN operation, to generate the final synthesized image, \(G(z; w)\), possessing the desired style and appearance characteristics. We train the model with our UV texture map dataset on 10000 UV maps from the FFHQ-UV dataset \cite{Bai_2023_CVPR}, aiming to generate UV maps of albedo textures. This allows the model to learn the distribution of UV maps. Post-training, we can sample points from the latent space and pass them through the model, generating UV maps similar to those of the dataset. When wrapped onto a mesh, these UV maps give rise to the appearance of a realistic human heads. 
\subsection{Interpolating semantic features}
\label{sec:interpolate}

GANs learn various semantics in linear subspaces of the latent space, which can be manipulated without retraining the model. Building upon the idea by \cite{shen2020interfacegan}, we train a linear Support Vector Machine (SVM) \cite{cristianini2000introduction} to classify between the levels of each semantic features within the learned latent space. For our model, we choose the criteria of age, facial hair and gender, as per the available labels from the FFHQ-UV dataset. However, it must be noted that this method can be scaled to any number of semantic features given a labeled dataset. Using the labels, the SVM learns to classify the latent vectors, treating those with highest attribute values in the available labels as positive samples, and those with the lowest values as negative. After training, the coefficients of the SVM are used as the boundary vector for the features. To perform a modification of a semantic in a given image with N steps, we use a projection method proposed in \cite{karras2020training} to get the latent vector from the image, and subsequently interpolate N steps in the latent space along the direction orthogonal to the boundary vector. These interpolated steps correspond to incremental manipulation in the semantic features.  By ensuring orthogonalization of all the boundaries, we disentangle the interpolation of semantic features. The ability to linearly interpolate with desired number of steps allows precise control over the manipulation, crucial for the process of designing and manipulating appearance of 3D human head models. 

\section{Experiment}
\label{sec:experiment}

We select 10000 UV maps from the FFHQ-UV dataset to train our model. We choose the architecture of \cite{karras2020training} since it is the state-of-the-art method for generating images resembling the training data. We train it on the dataset for 3000 epochs on NVIDIA GeForce RTX 2080 Ti. Each epoch was run for 4000 images, following the official PyTorch StyleGAN ADA implementation\footnote{\url{https://github.com/NVlabs/stylegan2-ada-pytorch}}.

Then we use the labels from the dataset to train a SVM, to be able to classify the latent vectors as per the attributes, as mentioned in Sec \ref{sec:interpolate}. 





\section{Results}
\label{sec:results}

\begin{figure}[t]
    \centering
    \includegraphics[width=0.6\textwidth]{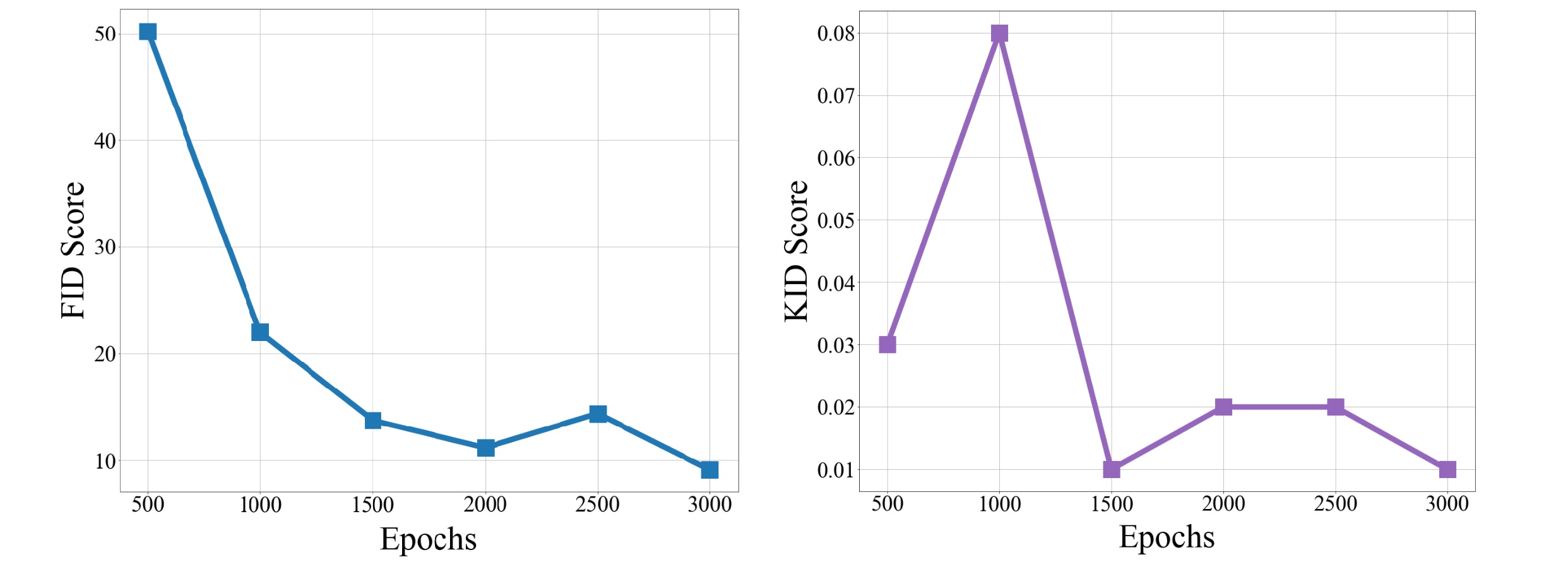}
    \caption{FID and KID scores over 3000 epochs: Low values of FID and KID scores indicate high-quality of generated samples}
    \label{fig:FID_KID}
\end{figure}

After training, we have a learned space of UV texture maps. We show the improvements while training the model using Fréchet Inception Distance (FID) score \cite{heusel2017gans} and Kernel Inception Distance (KID) \cite{binkowski2018demystifying} scores across the epochs (Fig.  \ref{fig:FID_KID}, both of which measure the similarity between real and generated samples. Lower values of FID and KID indicate that the generated samples are of high quality, closely resembling the real samples in terms of visual appearance.

\subsection{Interpolated results}

To visualize the working of SemUV on head UV texture maps, we sample from the dataset and perform semantic manipulation on 3 the features: age, facial hair, and gender.

\begin{figure}[h!]
     \begin{subfigure}[b]{0.3\textwidth}
         \centering
         \includegraphics[width=\textwidth]{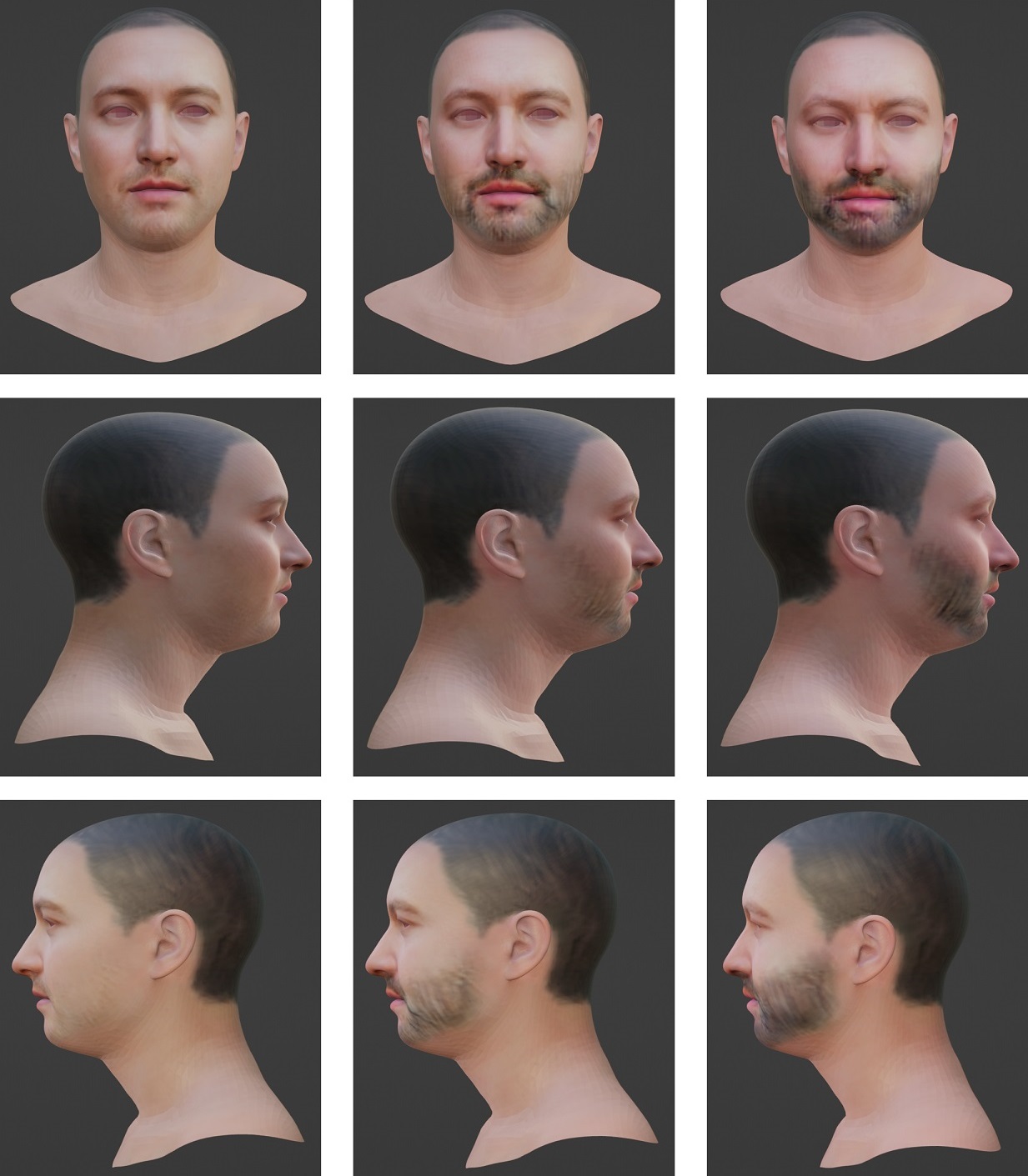}
         \caption{Facial hair}
         \label{fig:beard}
     \end{subfigure}
     \hfill
     \begin{subfigure}[b]{0.3\textwidth}
         \centering
         \includegraphics[width=\textwidth]{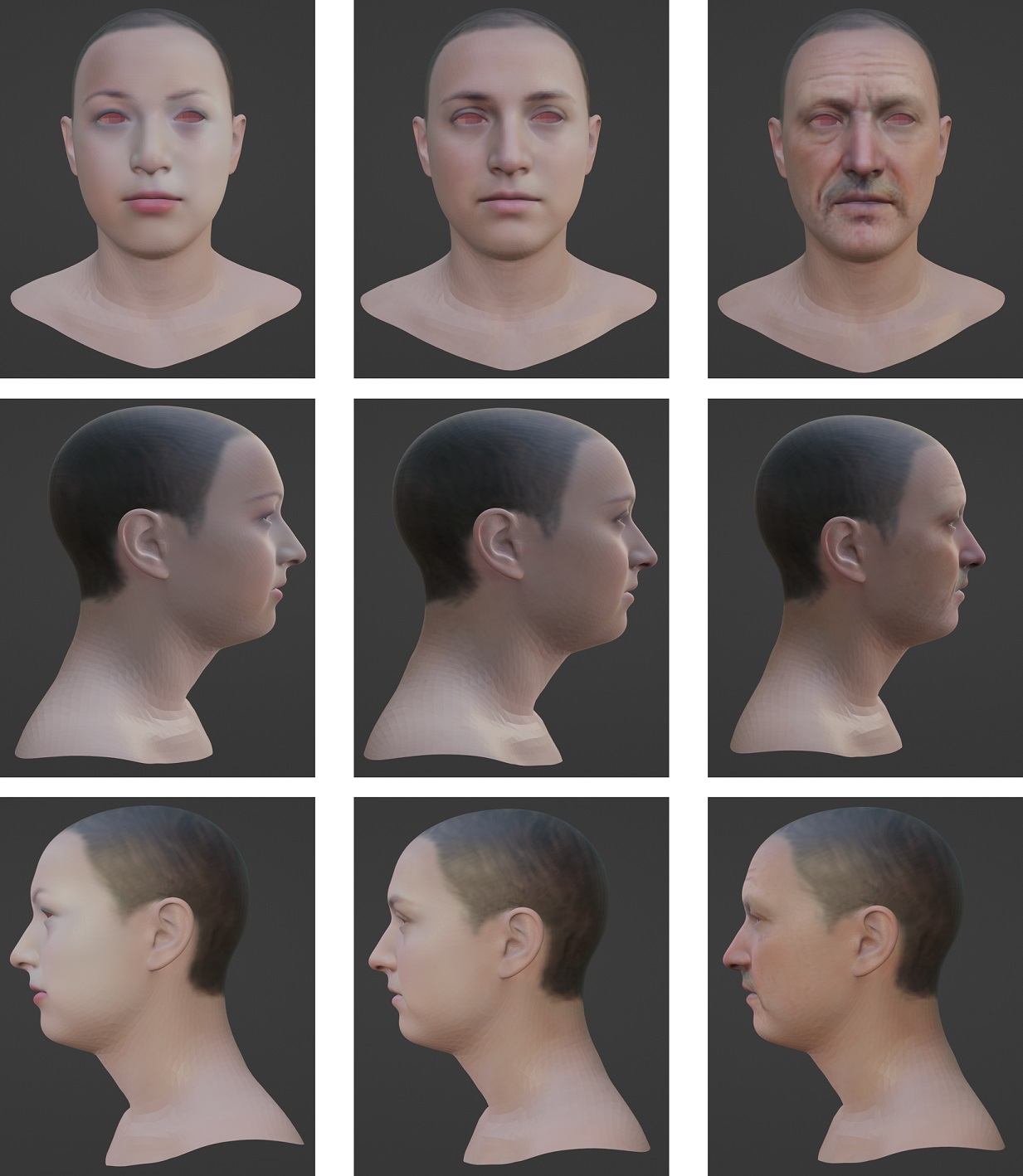}
         \caption{Age}
         \label{fig:age}
     \end{subfigure}
     \hfill
     \begin{subfigure}[b]{0.3\textwidth}
         \centering
         \includegraphics[width=\textwidth]{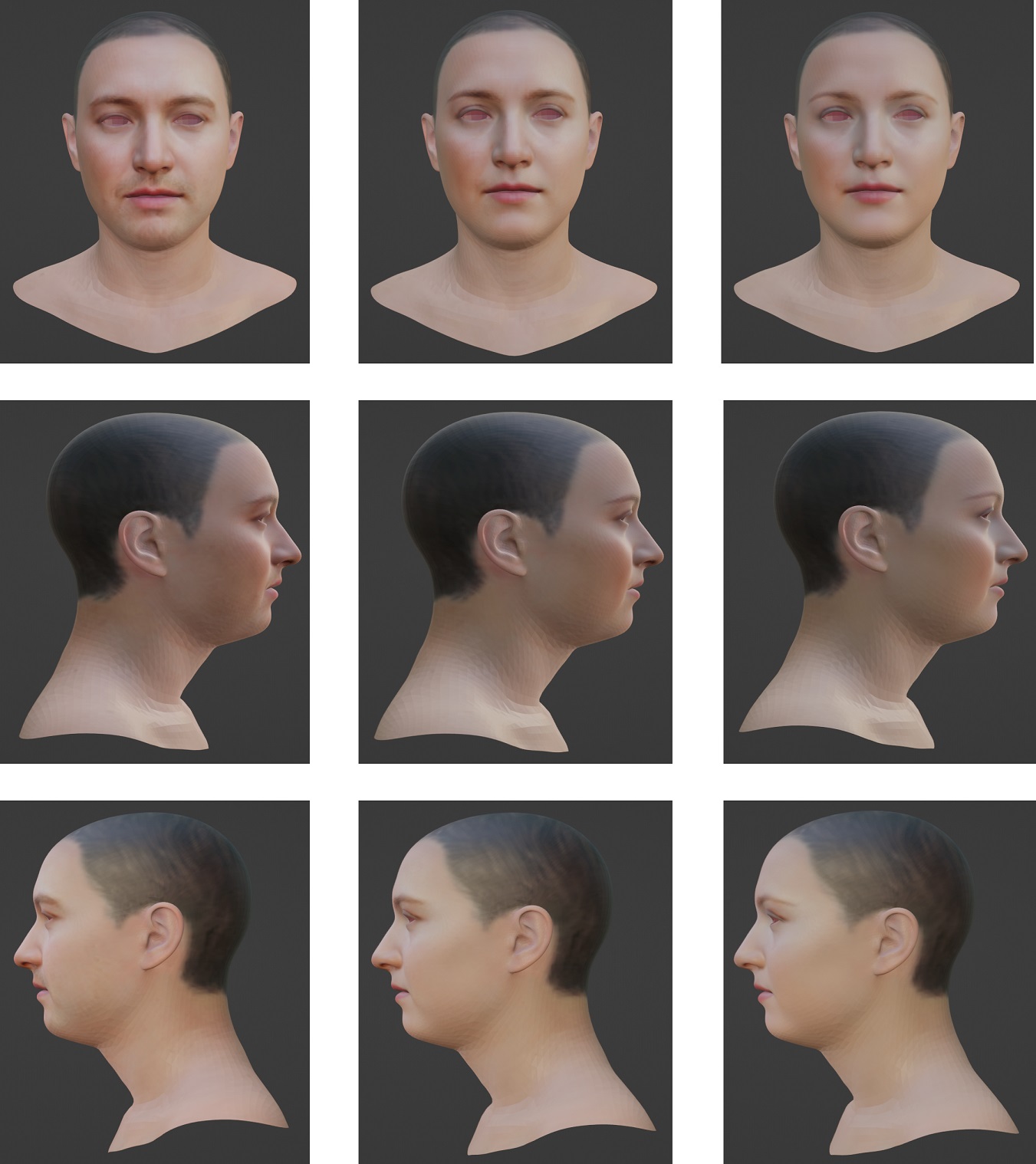}
         \caption{Gender}
         \label{fig:gender}
     \end{subfigure}
        \caption{Semantic manipulation results from (top) front, (middle) right and (bottom) left viewpoint. The changes in the features are made from left to right.}
        \label{fig:manipulations}
\end{figure}

After performing the manipulations, we project the texture maps on the 3D mesh provided by the dataset on Blender \footnote{\url{https://www.blender.org/}}, a 3D graphics engine. Once we have projected the UV map, we have a complete 3D head model which we can view from any direction. Figs. \ref{fig:beard}, \ref{fig:age}, \ref{fig:gender} show the result of the semantic manipulations on the head figure from various angles. 

\noindent \textbf{Age:} For the change in age in Fig. \ref{fig:age}, going from left to right we observe the features of the face being similar to that of a child, young teen, and a grown man respectively. There is change in the amount of facial hair such as moustache, beard and eyebrows. Also with age, there is an increase of high texture features such as wrinkles.

\noindent \textbf{Gender:} For the change in the gender in Fig. \ref{fig:gender}, we can observe the difference in face features such as lip shape, eyelid shape and eyebrow shape. Overall, female \textit{(right)} has sharper features and smoother skin. The male \textit{(left)} version had slight facial hair around the mouth area, a feature implicitly learned by the generative model.

\noindent \textbf{Facial hair:} For the change in facial hair in Fig. \ref{fig:beard}, we observe the gradual growth of hair as we go from left to right. We note that the growth of hair is same as that of humans, i.e. first slight growth of moustache and beard around the mouth region, followed by more dense beards in the sides of the face. 

Overall, we observe that the identity of the figure has been preserved in all the cases, and high level features are visible on the faces. Also, the changes are equally visible from all the angles such as front \textit{(top)}, left \textit{(middle)} and right \textit{(bottom)}, ensuring a structurally consistent editing of semantic features.

\subsection{Comparing with image-space manipulations}

\begin{figure}[h!]
    \centering
    \includegraphics[width=\textwidth]{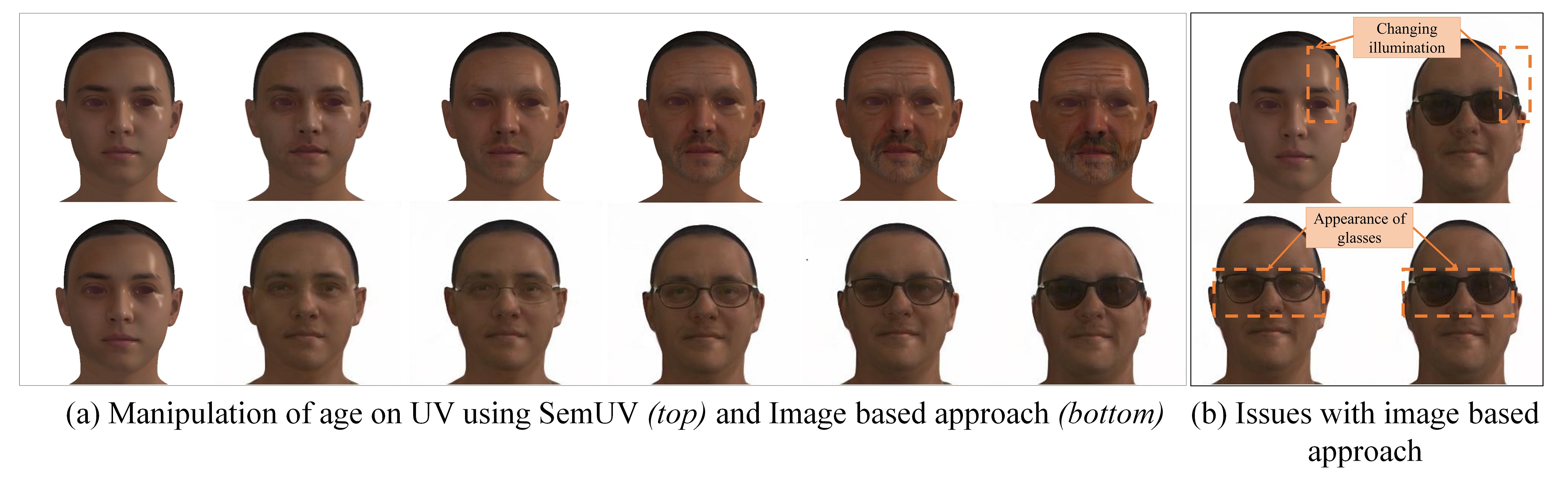}
    \caption{\textbf{(a)} Results of SemUV \textit{(top)} vs Image based approach \textit{(bottom)} \textbf{(b)} As we can see, changes in the image space can result in unwanted changes, change in identity, change in illumination, as well as image level artefacts. \textbf{SemUV}, is successfully able to add age related features without altering other aspects of appearance.}
    \label{fig:result_UV_vs_IMG}
\end{figure}

To evaluate the performance with respect to 2D image based approaches, we select the texture map obtained using the pipeline of FFHQ-UV on a custom image, and manipulate the age of the virtual head. 

We choose age attribute in particular because it is a common attribute between the FFHQ-UV labels, and InterFaceGAN \cite{shen2020interfacegan}, which we use as a benchmark for performing semantic manipulation after inverting a given face image using \cite{zhu2020domain}. Gender was another common attribute between the two approaches, however the presence of hair in the image space, and lack of hair on the 3D head could give biased results while user evaluation.

We project the base texture map, along with the manipulated ones on the generic head mesh provided in the official repository of FFHQ-UV. We use Pytorch3D to render the head display with the given mesh and textures, with a field of view of 20 degrees to approximate the view in face-images datasets. From the base image, we use \cite{zhu2020domain} to get a latent vector, and then interpolate the age feature using the official implementation. 

\noindent \textbf{Qualitative analysis}
The outputs obtained through SemUV contain appropriate wrinkles, signifying increasing age, along with some white facial beard. We can see the that on the face of the transformed image, glasses appear across interpolated results. 

\noindent \textbf{Computational comparison} From the figure, we can see that for manipulating a UV texture map using an image based method, there are extra steps for 1. Wrapping the UV map on a head mesh, 2. Rendering a display for the head, giving a head image, and 3. After manipulation, getting the unwrapped UV map of the face using FFHQ-UV pipeline. Considering these steps and ignoring the common tasks of GAN inversion to project into latent space, interpolating in the latent space, and also synthesizing manipulated image from latent space, we record the time taken for a set of 30 images to go through the pipeline (without considering the system operations taken by the CPU). We observe an additional time of \textit{15 minutes} taken by the image-space pipeline, which is primarily by the step of generating a UV map from a given image. This makes the method of SemUV faster than image based operation, allowing it to scale with multiple images and making it more suitable for real-time appearance editing in graphics based applications.

\begin{table}[b]
    \centering
  \caption{User evaluation results}
  \label{table:user_eval}
  \begin{tabular}{|c|c|c|}
  
    \hline
    \textbf{Question no} & \textbf{Choices for our approach}  & \textbf{p-value} \\
    \hline
    \textbf{Q1} & 86.67 \%   & 0.000030    \\
    \textbf{Q2} & 93.34 \%   & 0.000000    \\
    \textbf{Q3} & 86.67 \%   & 0.000030    \\
    \textbf{Q4} & 86.67 \%   & 0.000030    \\
    \textbf{Q5} & 73.34 \%   & 0.008062    \\
    \textbf{Q6} & 80.00 \%   & 0.000715    \\
    \textbf{Q7} & 86.67 \%   & 0.000030    \\
    \textbf{Q8} & 90.00 \%   & 0.000004    \\

    \hline
    
\end{tabular}
\end{table}

\noindent \textbf{User evaluation} For evaluating the performance of our approach as compared to the image based manipulations, a set of subjects were asked 8 questions as follows:

\noindent Q1. Which sequence (left to right) is better at showing increasing age?

\noindent Q2. Which sequence (left to right) is better at ensuring ONLY age is changed and no other aspect of appearance is changed?

\noindent Q3. Which sequence (left to right) better ensures lighting/illumination is maintained?

\noindent Q4. \& Q5. In which pair of images does the identity and facial structure of the left image match the most with that of the right image? (without considering age change)

\noindent Q6. \& Q7. In which pair of images does the expression of the left image match the most with that of the right image?

\noindent Q8. Out of the given sequences A and B, in which one does the changes in head pose seem most smooth and structurally consistent?

\begin{figure}[t]
    \centering
    \includegraphics[width=\textwidth]{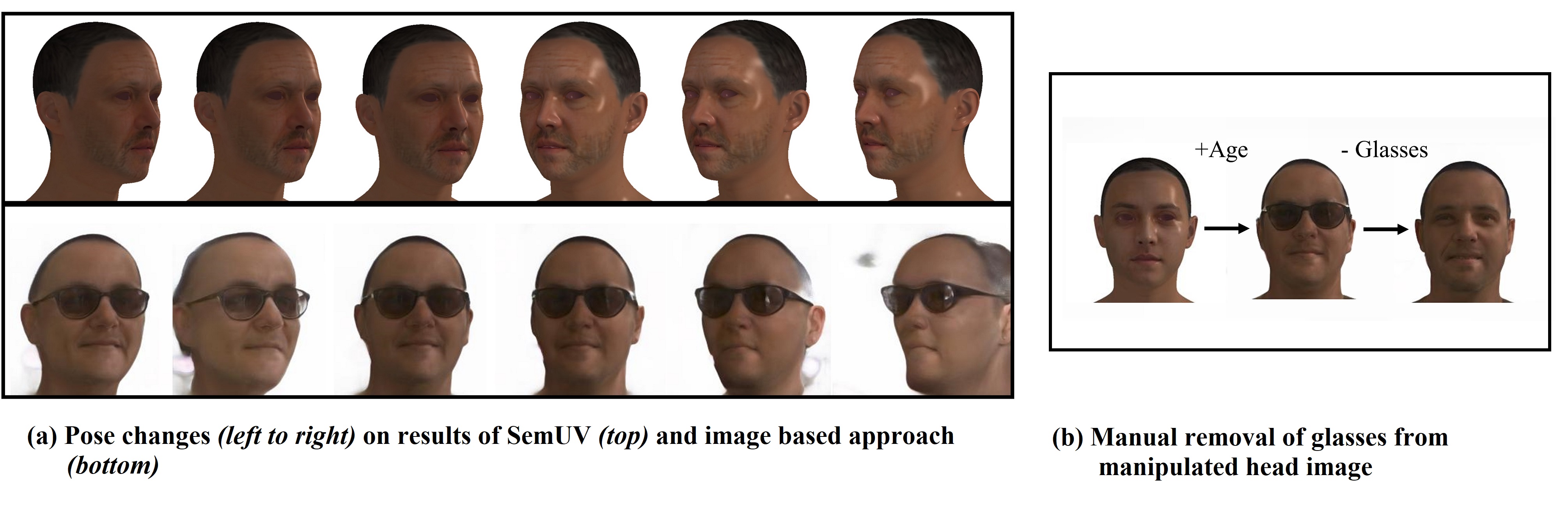}
    \caption{(a) Pose changes \textit{(from left to right)} on results of SemUV and image based approach (b) Removing glasses using \cite{shen2020interfacegan}}
    
    \label{fig:result_poses_glasses}
\end{figure}

For Q1, Q2, Q3, participants were shown Fig. \ref{fig:result_UV_vs_IMG} (a), i.e. two sequences of manipulation of age, one using SemUV, and the other using image based approach. For Q4 and Q6, participants were shown the left-most and right-most images of Fig. \ref{fig:result_UV_vs_IMG} (a), i.e. pairs of images before and after manipulation of features. For Q5 and Q7, we performed an additional task of removing the glasses using  \cite{shen2020interfacegan} once again on the obtained final image, as we can see in Fig. \ref{fig:result_poses_glasses} (b). The idea was to see whether the aspects of identity and emotion are preserved after removing the glasses which, even though requires an additional step, would let us know how our approach compares without the addition of unwanted glasses. Finally, for Q8, participants were shown Fig. \ref{fig:result_poses_glasses} (a), i.e. two sequences of images with changing pose, the bottom one being changed using image based manipulation, and top ones being changed simply by changing the camera angle during rendering.

To determine if it performed significantly better than chance, we performed one-sided binomial tests at a 5 \% significance level ($\alpha = 0.05$) for each question, with N = 30 trials and K successes where K is the number of participants who preferred the results of our approach. Our null hypothesis (H0) is that \textit{our approach is not significantly better than chance}, while our alternative hypothesis (H1) is that \textit{our approach is preferred more often}.

There were 30 participants (18 male and 12 female) for the survey. The mean age of the participants is 26.93, with a standard deviation of 5.02. From Table \ref{table:user_eval} we can see that there is a significant observation that our approach is able to perform age manipulation better \textit{(Q1)}, by ensuring no other aspect of appearance is changed \textit{(Q2)} and maintaining lighting and illumination \textit{(Q3)}. Our approach also ensures that the identity \textit{(Q4 and Q5)} and expressions \textit{(Q6 and Q7)} are maintained better by our approach. Interestingly, this holds true  even after manually removing the glasses \textit{(Q5 and Q7)} which appeared while manipulating the age. Also, due to the inherent 3D nature of our approach, we see that head changes in pose is smooth and consistent \textit{(Q8)} in our approach. While this is a default advantage of any 3D approach, the result indicates that using image based pose changes could result in inconsistencies in the head, leading to unwanted outputs both in the image space, and also UV maps derived from them. 

To ensure uniform and spatially consistent appearance manipulation across the whole head surface, our approach proves effective due to its operation over the complete head albedo space.


\section{Discussion}
In this section, we look into the key insights regarding the importance of SemUV, based on our conducted experiments. We examine the limitations of our approach and explore potential future research directions aimed at enhancing our method and expanding the range of successful semantic manipulations of virtual heads in UV space.
 
\subsection{Significance of SemUV: }Having a model capable of performing manipulations directly in UV space presents a significant advantage for designers operating in the realm of 3D graphics. While transformations generated by models operating in the 2D image space, such as increasing hair length while changing gender from male to female or adding glasses when increasing age, may seem logically valid, they are not suitable for the process of graphic designing processes, where such elements are typically modelled separately. Moreover, employing image-based approaches for UV editing is time taking due considering the additional steps of conversions between UV and image space. Also, these additional steps can generate unwanted changes in semantic aspects of appearance, including identity, and result in artifacts and structural inconsistencies. In contrast, manipulating in UV space allows the generative model to focus exclusively on the high resolution (1024 $\times $ 1024) texture for head area, and be agnostic to other physical aspects such as structure, lighting and viewing direction. These advantages make SemUV significantly valuable in the area of virtual head modelling using 3D computer graphics.

\subsection{Limitations: }
While our approach is successfully able to perform manipulations in the semantic aspects, there are challenges in disentangling certain features in the latent space where the labels are noisy. In Fig. \ref{fig:SemUV_outputs}, we see that although there was no facial hair in the initial image, the aged version has added slight patches of white hair. Such challenges could be tackled by using more labelled examples, as well as modified methods for non-linear separation boundaries for latent space disentanglement.

\subsection{Future scope: } Some future research directions could explore various realistic face UV datasets. With a dataset of high quality and diverse representations of texture maps, more photorealistic images could be created which will result in even more realistic faces that will be perceptually indifferent from real faces, reaching the photorealistic quality of modern state-of-the-art generative models for 2D face images. This will further reduce the gap between 2D and 3D face manipulations. Apart from the quality of the images in the dataset, there could also be more sets of labels and attributes which correspond to other high level semantic features. These labels could be used for training our proposed pipeline so that there is even more control for variations in the human heads. Another possible future direction for the dataset could be to include normal maps, which include information about the depth of images. Representing details in skin requires depth information along with color. To accurately change features such as age, such details and textures in skin needs to be represented. Allowing the model to learn the normal maps along with albedo maps could enable better semantic manipulations and much higher quality of rendering of the human faces. Considering that these virtual humans are made keeping in mind a diverse range of demographics, the ability to perform manipulations of features such as race and gender will be helpful in introducing diversity in the virtual humans.

\section{Conclusion}

In this work, we have proposed SemUV: a novel, simple and effective deep learning based method for performing semantic manipulation of 3D human faces in UV texture maps. Leveraging a learned UV texture space from state-of-the-art high resolution dataset, we have successfully demonstrated manipulations of semantic features such as age, beard, and gender. We have conducted a user study to explore the perceived differences between our approach and traditional 2D image-based methods, supported by statistical tests to infer valuable insights for our method. Furthermore, we have highlighted the potential real-life applications of successful semantic manipulation in the texture space, including VR and AR. By enabling more efficient and resource-effective design processes for realistic virtual humans, our approach promises to significantly benefit 3D graphic designers across various domains. This work contributes a promising solution to the ongoing challenges in 3D face manipulation, offering both practical applications and valuable insights for future research advancements in the field of virtual head modelling.

\section*{\small Acknowledgment}
\vspace{-0.3cm}
\small This project was funded by MINRO center at IIITB.


\bibliography{main}
\bibliographystyle{splncs04}


\end{document}